\documentclass[11pt, oneside]{article}   	% use "amsart" instead of "article" for AMSLaTeX format
\usepackage{geometry}                		% See geometry.pdf to learn the layout options. There are lots.
\usepackage{graphicx}				% Use pdf, png, jpg, or eps with pdflatex; use eps in DVI mode
\usepackage{amssymb}

\usepackage{subfigure}
\usepackage{amsmath}

\title{An Approach to the Analysis of the South Slavic Medieval Labels Using Image Texture}
\author{Darko Brodi\'c\footnote{University of Belgrade, Technical Faculty in Bor, V.J. 12, 19210 Bor, Serbia, dbrodic@tf.bor.ac.rs}, Alessia Amelio\footnote{Institute for High Performance Computing and Networking, National Research Council of Italy, CNR-ICAR, Via P. Bucci 41C, 87036 Rende (CS), Italy, amelio@icar.cnr.it}, Zoran N. Milivojevi\'c\footnote{College of Applied Technical Sciences, Aleksandra Medvedeva 20, 18000 Ni\v s, Serbia, zoran.milivojevic@vtsnis.edu.rs}}
\date{}							% Activate to display a given date or no date

\begin{document}
\maketitle

\begin{center}
\bf Abstract
\end{center}
{\scriptsize \bf The paper presents a new script classification method for the discrimination of the South Slavic medieval labels. It consists in the textural analysis of the script types. In the first step, each letter is coded by the equivalent script type, which is defined by its typographical features. Obtained coded text is subjected to the run-length statistical analysis and to the adjacent local binary pattern analysis in order to extract the features. The result shows a diversity between the extracted features of the scripts, which makes the feature classification more effective. It is the basis for the classification process of the script identification by using an extension of a state-of-the-art approach for document clustering. The proposed method is evaluated on an example of hand-engraved in stone and hand-printed in paper labels in old Cyrillic, angular and round Glagolitic. Experiments demonstrate very positive results, which prove the effectiveness of the proposed method.\\\\
{\bf Keywords:} Classification, Historical document, Optical character recognition, Pattern recognition, Script identification, Medieval labels.}

\section{Introduction}

Natural scene images include short text information with their background. The most important part of these scenes represents the extraction of the text information. This process typically includes following stages: text detection, text localization, text extraction and enhancement and recognition \cite{Zhang2013}. The recognition of the medieval labels is even more complicated due to the presence of the high noise elements. Furthermore, the process of the script recognition in the South Slavic medieval labels is questionable due to variety of scripts included.

Old Slavonic is the oldest documented Slavic language dated from 850 - 1100 AD. In the middle of the 9th century Byzantine Greek missionaries St. Cyril and St. Methodius standardized the Slavic language as a task given by the Byzantine emperor. They used it to translate the Bible and related liturgical documents from Greek. It was a part of the attempt to Christianize the Slavic people, because they were pagans. Old Church Slavonic was based upon the dialect of Slavic people living around Thessalonika. Furthermore, St. Cyril and St. Methodius created a new alphabet, which was accustomed to the Old Church Slavonic language. It was called Glagolitsa (round Glagolitic script), which represents the oldest known Slavic alphabet. During their mission to Great Moravia, they spread the Glagolitic script to Bosnia, Croatia, Serbia, Lesser Poland. Later, their students brought the Glagolitic alphabet to Bulgaria. Glagolitic alphabet has 41 letters. However, it was replaced by Cyrillic alphabet created in Preslav schools by Cyril's students. Initial Cyrillic alphabet has 44 letters. However, due to its origin to Greek uncial script, Cyrillic alphabet was suited to the writing of Old Church Slavonic. It was valid because it follows the principle that one letter represents one significant sound. In Croatia, the Glagolitic script evolved into its new version called angular (angular) Glagolitic script,  which was used up to the 19th century.

South Slavic medieval labels and graffiti \cite{Fucic82} represent the full elements or excerpts from hand-printed or stone engraved documents representing text objects. They were written by round Glagolitic, angular Glagolic and old Cyrillic scripts. The problem exists with hand-printed documents, because they used the writing rules that were changed in different historical periods, and from the regions where they were created. The situation is similar to the hand crafted stone engraved labels. Hence, the recognition of script and writing rules can clearly identify the historical context of the analyzed document.

The aim of this paper is to present an algorithm that can successfully identify the script of the South Slavic medieval labels. Previously, a similar approach was proposed in refs. \cite{BrodicSC}, \cite{Brodic14} in order to differentiate the South Slavic scripts. The problem persists with medieval labels, because they consist of a very small number of characters and incorporate the change of writing rules typical for certain historical periods. Up to now, there is no such attempt to solve a similar problem. Furthermore, these historical documents are mainly in bad condition, which includes the ink noise in paper or noise in stones. Still, the proposed algorithm is relatively prone to errors, which leads to good script identification results. Hence, it represents a state of the art  approach.

The paper is organized as follows. Section 2 describes the proposed algorithm. Section 3 illustrates the experiment. Section 4 presents and discusses the results. Section 5 gives the conclusions.

\section{The Algorithm}

\subsection{The Flow of the Proposed Algorithm}
The aim of the proposed algorithm is the identification of different scripts used in South Slavic medieval labels. It accomplishes this task in multi-stage procedure. The algorithm consists of the following stages: (i) horizontal projection profile of the text to distinct different text lines, (ii) distribution of the blob heights and its center point, (iii) classification according to typographical features, (iv) mapping according to script types, (v) creating an equivalent image pattern, (vi) features extraction of the image pattern according to the texture analysis, (vii) classification of the extracted features.

\subsection{Typographical Features}

\begin{figure}[!ht]
\begin{center}
\includegraphics[width=11cm, keepaspectratio]{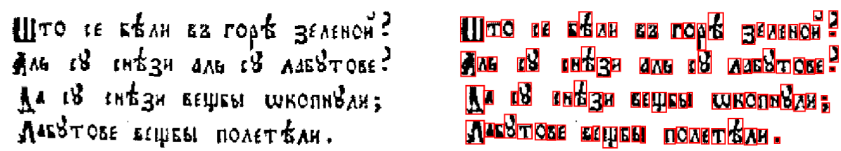}
\caption{Text segmentation, from left to right: initial text (written in the old Cyrillic script), bounding box operation around each blob in the text. }
\label{Fig1}
\end{center}
\end{figure}

At the beginning, we suppose that the skew identification and correction are previously carried out. Hence, it is out of the scope of the proposed algorithm. The first stage of the proposed algorithm represents the extraction of typographical features from the text image. It includes three aforementioned steps given as a horizontal projection profile of the text to distinct different text lines, distribution of the blob heights and its center point, and its classification according to typographical features. Hence, the algorithm starts with the horizontal projection profiles, which define the text line position. Then, the blob extraction will be classified according to the previous text line segmentation. Fig. \ref{Fig1} shows the initial text as well as its blob (letter) segmentation by bounding box extraction procedure. Then, each bounding box is used to extract its height and center point \cite{Preparata95}.

\begin{figure}[!ht]
\begin{center}
\includegraphics[width=7cm,height=7cm,keepaspectratio]{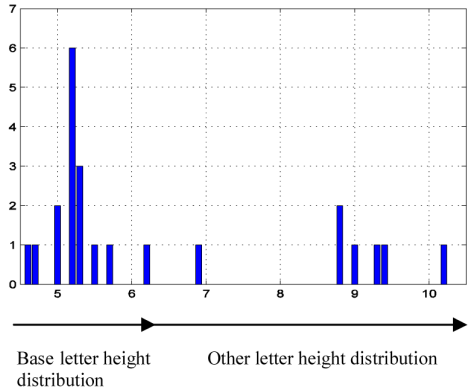}
\caption{Distribution of the character heights.}
\label{Fig2}
\end{center}
\end{figure}

Fig. \ref{Fig2} shows the character heights distribution given by the bounding boxes.

\begin{figure}[!ht]
\begin{center}
\includegraphics[width=10cm,height=10cm,keepaspectratio]{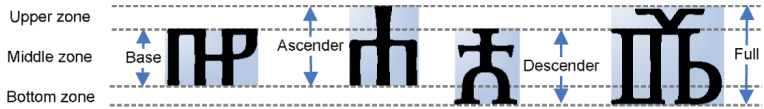}
\caption{Grouping of the letters according to typographical feature in the text line.}
\label{Fig3}
\end{center}
\end{figure}

According to the typographical features, each letter can be divided into the following letter groups \cite{Zram98}: (i) base letter, (ii) ascender letter, (iii) descendent letter, and (iv) full letter. Fig. \ref{Fig3} illustrates these groups of letters.

\begin{figure}[!ht]
\begin{center}
\includegraphics[width=10cm,height=10cm,keepaspectratio]{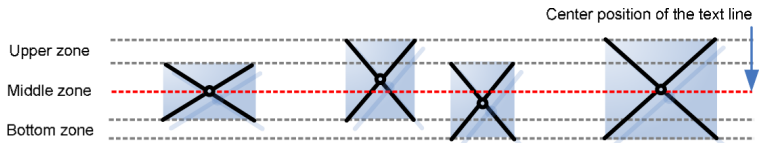}
\caption{Center point position of different character types (script types) in the text line.}
\label{Fig4}
\end{center}
\end{figure}

From Fig. \ref{Fig3}, it is easy to realize that there exist three different letter heights. Obviously, the base letters have the smallest heights, while the full letters have the tallest heights. Furthermore, the ascender and descendent letters characterize similar heights in between the base and full letters.
Fig. \ref{Fig4} illustrates the center point of different script types.

Different script types can be recognized by their different center positions. Although, the base and full letter have the similar position of the center point, i.e. close to the center position of the text line, they can be recognized by different heights. Furthermore, ascender letter has the center point in the upper part of the text line, while the descendent letter has the center point in the lower part of the text line. Hence, they can differentiate easily in spite of its similar heights.

\subsection{Mapping Script Types}

\begin{figure}[!ht]
\begin{center}
\includegraphics[width=7cm,height=7cm,keepaspectratio]{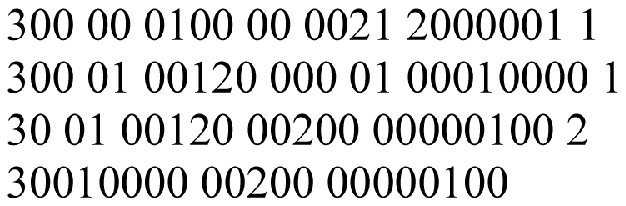}
\caption{Mapping of the initial text to coded text.}
\label{Fig5}
\end{center}
\end{figure}

Second stage of the algorithm represents the mapping according to script types and  creating an equivalent image pattern. Hence, the obtained division of letters to base, ascender, descendent and full letters is mapped into the coded text which corresponds to image texture. In this way, the following mapping is carried out: base letter to 0, ascender letter to 1, descendent letter to 2, and full letter to 3 \cite{BrodicSC}. According to proposed mapping, the initial text is converted to coded text given in Fig. \ref{Fig5}.

\begin{figure}[!ht]
\begin{center}
\subfigure{
\includegraphics[width=10cm,height=10cm,keepaspectratio]{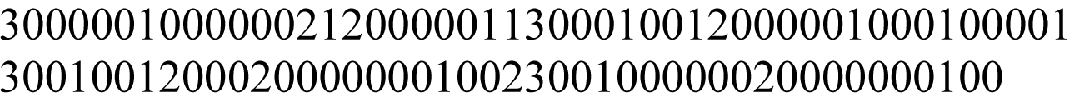}
}\\
\subfigure{
\includegraphics[width=10cm,height=10cm,keepaspectratio]{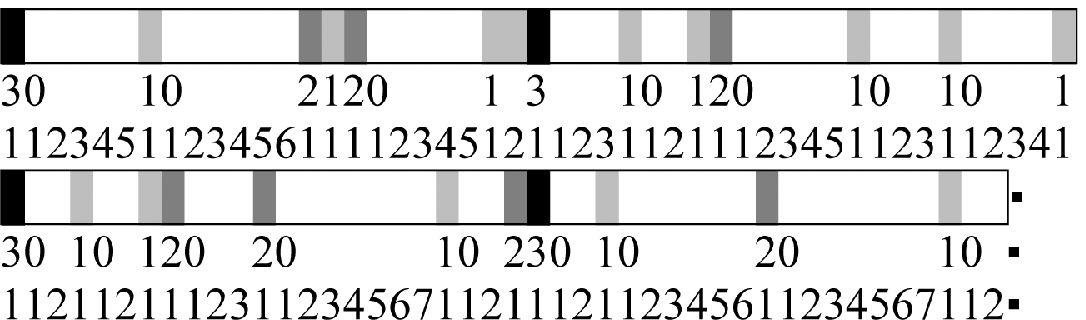}
}
\caption{Mapping of the script types to the corresponding image pattern, from top to bottom: coded text, process of mapping to gray levels of the image, identification of different levels of gray and the numbers that represent the repeating number of each gray level ($\blacksquare$ marks the end of the text, i.e. image pattern).}
\label{Fig6}
\end{center}
\end{figure}

The given coded text can be mapped into the image. Each text line is continued in the next line up to the end of the text. Furthermore, each of four different numbers corresponds to different level of gray in the image. Accordingly, the image is given as 1-D image with four gray levels. Fig. \ref{Fig6} illustrates the procedure of creating an image pattern from the coded text.

\subsection{Feature Extraction of the Texture}
Texture is a measure of the intensity variation in the image surface \cite{Tan98}. Hence, it is suitable for information extraction, which can be used to quantify the properties like image smoothness, coarseness, and regularity. Accordingly, the texture can be used to calculate statistical measures of the image.

\subsubsection{Run-length Statistics}
Run-length statistical analysis is used to extract texture features and to quantify texture coarseness \cite{Galloway75}. A run represents a set of consecutive pixels characterized with the same gray-level intensity in a specific direction of the texture. Hence, the fine textures characterize the long runs, while coarse textures contain short runs.

\begin{figure}[!ht]
\begin{center}
\includegraphics[width=6cm,height=6cm,keepaspectratio]{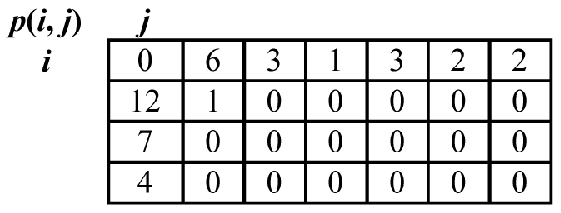}
\caption{Run-length matrix p(i, j) obtained from the coded text given in Fig. \ref{Fig6}.}
\label{Fig7}
\end{center}
\end{figure}

Let's suppose that we have an image I featuring X rows, Y columns and L levels of gray intensity. First, the run-length matrix p(i, j) is extracted. It is defined by specifying direction and then counting the occurrence of runs for each gray level and length in this direction. Its number of rows corresponds to M, i.e. the maximum number of gray levels, while the number of columns corresponds to N, i.e. to the maximum run length. In our case, each element of the run-length matrix p(i, j) represents the gray level run-length of image I (1-D matrix) that gives the total number of occurrences of gray-level runs of length $j$ and of the gray-level intensity value $i$. Accordingly, a set of consecutive pixels with identical intensity values constitutes a gray level run.  Fig. \ref{Fig7} shows p(i, j) obtained from the coded text presented in Fig. \ref{Fig6}.

The extraction of various texture features from the run-length matrix is given as follows \cite{Galloway75}: (i) Short run emphasis (SRE), (ii) Long run emphasis (LRE), (iii) Gray-level non-uniformity (GLN), (iv) Run length non-uniformity (RLN), and (v) Run percentage (RP).

SRE measures the distribution of short runs. It is dependent on the occurrence of short runs. For fine textures, it is expected to receive large values. It is calculated as:
\begin{equation}
SRE = \frac{1}{n_r} \sum_{i=1}^{M} \sum_{j=1}^{N} \frac{p(i,j)}{j^2}.
\end{equation}

LRE measures the distribution of long runs. It is dependent on the occurrence of long runs. For coarse textures, it is expected to reach large values.  It is calculated as:
\begin{equation}
LRE = \frac{1}{n_r} \sum_{i=1}^{M} \sum_{j=1}^{N} p(i,j) \cdot j^2.
\end{equation}

GLN measures the similarity of gray level values throughout the image. If the gray level values are alike throughout the image, then it receives small values. It is calculated as:
\begin{equation}
GLN = \frac{1}{n_r} \sum_{i=1}^{M}( \sum_{j=1}^{N} p(i,j))^2.
\end{equation}

RLN measures the similarity of the length of runs throughout the image. If the run lengths are alike throughout the image, then it reaches small values. It is calculated as:
\begin{equation}
RLN = \frac{1}{n_r} \sum_{j=1}^{N}( \sum_{i=1}^{M} p(i,j))^2.
\end{equation}

RP measures the homogeneity and distribution of runs of an image in a specific direction. It receives the largest values when the length of runs is 1 for all gray levels in a specific direction. It is calculated as:
\begin{equation}
RP = \frac{n_r}{n_p}.
\end{equation}

In the above equations, $n_r$ represents the total number of gray-level runs, while $n_p$ is the number of pixels in the image I.

The extraction of texture features from the run-length matrix is extended by the following two measures \cite{Chu90}: (i) Low gray-level run emphasis (LGRE) and (ii) High gray-level run emphasis (HGRE).

LGRE is calculated as:
\begin{equation}
LGRE = \frac{1}{n_r} \sum_{i=1}^{M} \sum_{j=1}^{N} \frac{p(i,j)}{i^2}.
\end{equation}

HGRE is calculated as:
\begin{equation}
HGRE = \frac{1}{n_r} \sum_{i=1}^{M} \sum_{j=1}^{N} p(i,j) \cdot i^2.
\end{equation}

Additional four feature extraction functions gained by idea of joint statistical measure of gray level and run length are proposed in \cite{Dasar91}. They are: (i) Short run low gray-level emphasis (SRLGE), (ii) Short run high gray-level emphasis (SRHGE), (iii) Long run Low gray-level emphasis (LRLGE), and (iv) Long run high gray-level emphasis (LRHGE).

SRLGE is calculated as:
\begin{equation}
SRLGE = \frac{1}{n_r} \sum_{i=1}^{M} \sum_{j=1}^{N} \frac{p(i,j)}{i^2\cdot j^2}.
\end{equation}

SRHGE is calculated as:
\begin{equation}
SRHGE = \frac{1}{n_r} \sum_{i=1}^{M} \sum_{j=1}^{N} \frac{p(i,j) \cdot i^2} {j^2}.
\end{equation}

LRLGE is calculated as:
\begin{equation}
LRLGE = \frac{1}{n_r} \sum_{i=1}^{M} \sum_{j=1}^{N} \frac{p(i,j) \cdot j^2} {i^2}.
\end{equation}

LRHGE is calculated as:
\begin{equation}
LRHGE = \frac{1}{n_r} \sum_{i=1}^{M} \sum_{j=1}^{N} p(i,j) \cdot i^2 \cdot j^2.
\end{equation}

In this way, run-length statistical analysis extracts 11 feature measures.

\subsubsection{Adjacent Local Binary Pattern}
\begin{figure}[!ht]
\begin{center}
\includegraphics[width=11cm,height=11cm,keepaspectratio]{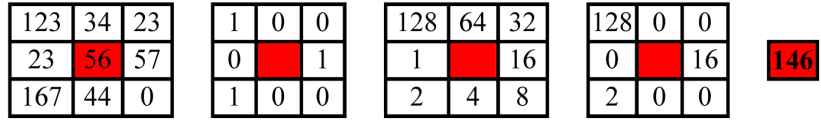}
\caption{Calculating LBP, from left to right: Initial window of interest with center point (56), Thresholding, Different powers of two depending on the pixel position, Calculation by multiplying the second one and the third one, and finally Sum of the LBP labels.}
\label{Fig8}
\end{center}
\end{figure}

Local binary pattern (LBP) represents the texture operator. It determines a magnitude relation between a center pixel and its neighbor pixels in some window of interest \cite{Ojala96}. Actually, it measures local image contrast variations.
LBP is calculated by thresholding the image intensity of the surrounding pixels with the intensity of the central pixel. Fig. \ref{Fig8} illustrates such a circumstance.

In this way, the pixels in the block are threaded by its center pixel value. If the intensities of neighboring pixels are higher than the center one, the corresponding pixel will be assigned as 1, otherwise it will be assigned as 0. Then, they are multiplied by different powers of two depending on the pixel position. At the end, they are summed in order to obtain a label for the center pixel. Typically, the  neighborhood of the center pixel includes eight pixels. Hence, we obtain $2^8 = 256$ different labels. For a given center pixel, the LBP is calculated as follows \cite{Ojala2002}:
\begin{equation}
LBP_{d,r}=\sum_{i=1}^{d}sign(I_i - I_{center}) \times 2^{i-1},
\end{equation}
where $d$ represents the number of neighboring pixels, while $r$ is the distance between the neighboring pixels and the center pixel. Accordingly, $I_{center}$ is the center pixel, while $I_i$ is the i-th neighboring pixel around $I_{center}$.  The sign is calculated as follows:

\begin{equation}
sign(x)=\begin{cases} 1, & \mbox{if }x\mbox{ $\ge 0$} \\ 0, & \mbox{if }x\mbox{ $< 0$}
\end{cases}.
\end{equation}

\begin{figure}[!ht]
\begin{center}
\includegraphics[width=12cm,height=12cm,keepaspectratio]{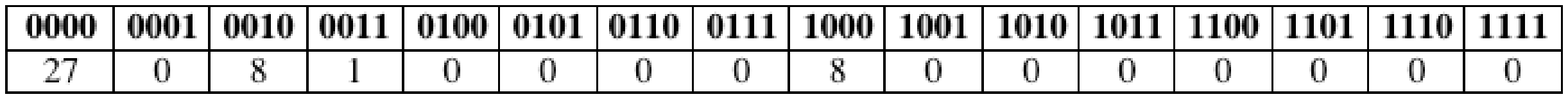}
\caption{ALBP obtained from the coded text given in Fig. \ref{Fig6}}
\label{Fig9}
\end{center}
\end{figure}

In our case, the coded text is given as a 1-D image. Hence, the neighborhood includes only 2 instead of 8 pixels. This leads to a total of $2^2 = 4$ different labels \cite{Brodic14}. The problem persists due to a too small number of extracted features for further analysis. Hence, some extension of the original LBP operator is eligible. Ref. \cite{Amelio14} introduces the combination of two LBP subsets. They represent LBP(+) and LBP($\times$), which are called adjacent local binary pattern (ALBP). LBP(+) considers four pixels, i.e. two adjacent pixels in horizontal and two adjacent pixels in vertical directions. LBP($\times$) examines the four diagonal pixels. In our case, we consider 1-D image, which corresponds only to horizontal direction. Hence, vertical and diagonal directions are insignificant. Accordingly, ALBP creates two adjacent LBP(+) establishing a 4-bit binary label leading to the total number of $2^4 = 16$ different labels. Fig. \ref{Fig9} shows ALBP (from '0000' to '1111') obtained from the coded text presented in Fig. \ref{Fig6}.
\vspace{-0.7cm}

\subsection{Feature Classification}
Documents in Cyrillic, angular and round Glagolitic scripts, codified by run-length and ALBP features, are discriminated by an extension of the Genetic Algorithms Image Clustering (GA-IC) method \cite{Amelio14}, called Genetic Algorithms Image Clustering for Document Analysis (GA-ICDA).
The base version GA-IC represents a genetic approach for clustering an image database, codified as a weighted graph. The nodes of the graph are the images, while the edges connect each node with its $h$-nearest neighbors. They represent the nodes associated with the $h$-lowest distances, in terms of similarity of the images corresponding to the nodes, with respect to the considered node. The $h$ parameter is related to the size of the neighborhood. The edges of the graph are weighted to represent the similarity among the nodes. Then, an evolutionary procedure is applied on this graph for clustering the nodes and obtaining the image classes.

GA-ICDA presents three important novelties making the evolutionary procedure suitable for text document clustering. They consist in the feature representation, in the  graph construction and in the cluster detection.

The first modification is in the feature representation. Each node of the graph is a  document in Cyrillic, angular or round Glagolitic scripts. It is represented as a feature vector composed of the 11 run-length statistical analysis values (5 features of Galloway \cite{Galloway75}, 2 features of Chu et al. \cite{Chu90}, and 4 features of Dasarathy and Holder \cite{Dasar91}) and of the 16 ALBP referent micro patterns from '0000' to '1111'.

The second novelty consists in modifying the graph construction by introducing a concept derived from that of Matrix Bandwidth \cite{Marti2001}. Specifically, consider $f$  to be the node ordering induced by the graph adjacency matrix, a one-to-one mapping from graph nodes to integer identifiers $f: V \to \{1, 2,.., n\}$. For each node $v \in V$ whose corresponding identifier is $f(v)$, we evaluate the difference between $f(v)$ and the identifiers $F$ of the nodes $nn^h_v$ in the $h$-neighborhood of $v$. Then, for each node $v$, we include edges between $v$ and the only nodes in $nn^h_v$ whose identifier difference $| f(v)- f(nn^h_v(i))|$ is less than a threshold value $T$.

Finally, a refinement phase is introduced at the end of the evolutionary procedure. For each detected cluster $c_a$, we find the cluster $c_b$  having the minimum distance from it. Then, we merge the pair of detected clusters $<c_a, c_b>$ and repeat the procedure until a fixed number of clusters is reached. The distance between two clusters is computed as the $L_1$ norm between the two farthest text document feature vectors, one for each cluster.

\section{Experiments}
The proposed method is evaluated on two custom-oriented databases of short documents (labels) written in old Cyrillic, angular and round Glagolitic scripts. Each of the documents has less than 100 characters.

The first database is composed of 15 labels, where 5 labels are in Cyrillic script, 5 labels are in angular Glagolitic and 5 labels are in round Glagolitic. Then, the first database is extended to create a second database of 20 labels. Specifically, a new set of 5 labels in angular Glagolitic is added to the original set. The 5 labels added to the angular Glagolitic set were established in the early historic period when the round Glagolitic script was in the process of conversion to the angular Glagolitic script. It represents the historic pre-period of the angular Glagolitic script.

\section{Results and Discussion}
In the following section, we assess the ability of the GA-ICDA classifier together with the new document feature coding in correctly distinguishing between the three different kinds of scripts in the databases. Consequently, the experiment consists in adopting GA-ICDA on the run-length and ALBP feature vectors of the labels in old Cyrillic, angular and round Glagolitic scripts. The goal of the experiment is to evaluate if the classifier can discriminate a label given in a particular kind of script (i.e. Cyrillic script) instead of another one (i.e. angular or round Glagolitic script). The classification task is particularly difficult in this context because we have old labels hand-engraved in stone and hand-printed on paper.

Also, in order to demonstrate the robustness of our technique, we present two kinds of tests at two different difficulty levels. In the first one, GA-ICDA classifies the first database of 5 labels in Cyrillic, 5 labels in angular Glagolitic and 5 labels in round Glagolitic. In the second test, GA-ICDA is evaluated on the second more complex database of 5 labels in Cyrillic, 10 labels in angular Glagolitic and 5 labels in round Glagolitic. The 5 labels added to the angular Glagolitic class make the classification process much more difficult than for the first database, because the real distinction between the three kinds of scripts begins much more complex.

We use a trial and error procedure for tuning the parameters of GA-ICDA. It is adopted on benchmark documents for computing the parameters giving the best classification results. Then,  the obtained parameter values are used for clustering of the scripts in the custom-oriented databases. Consequently, we fix the $h$ value of the node neighborhood to 15 and the $T$ threshold value to 4, for the first database in the first test, and the $h$ value of the node neighborhood to 20 and the $T$ threshold value to 5, for the second database in the second test.
For the performance evaluation of our technique, we compare GA-ICDA with other two unsupervised classifiers, specifically with the Average Linkage Hierarchical clustering and the K-Means method,  both well-known approaches also adopted for text classification \cite{Yuyu13}, \cite{Aggar12}, \cite{Zhu2006}. Hierarchical and K-Means classifiers use the same run-length and ALBP feature representation as GA-ICDA.

Precision, recall and f-measure are the performance measures adopted for the evaluation of the classifiers. They are calculated for each script class with respect to the ground-truth (true) partitioning of the scripts in the three classes by using the confusion matrix \cite{Andrews2007}, \cite{Vries12}. Also, because the association of the found clusters with the true class is required for computing precision and recall, and it is not given a priori when clustering is computed, we assume that each cluster corresponds to the real script class which is in majority in that cluster. Also, Normalized Mutual Information (NMI) is used for computing the similarity of the found clustering with the ground-truth (true) division of the scripts \cite{Andrews2007}, \cite{Mann2009}. This measure ranges between 0 (total mismatch) and 1 (complete match). Consequently, such a measure is used as a criterion for evaluating the goodness of the partitioning found from the algorithm and for comparing the found clustering solution with the solution of the different approaches. As NMI is a measure evaluating the similarity of the overall partitioning in different scripts, one single value is reported for Cyrillic, angular and round Glagolitic classes.

Experiments have been run on a Desktop computer quad-core 2.6 GHz with 8 Gbyte of RAM and Windows 8. Table 1 reports the results of GA-ICDA classification on the first database compared with hierarchical and K-Means (first test). Table 2 illustrates the results of the GA-ICDA, hierarchical and K-Means classifiers on the second database (second test). In bold are the cases when GA-ICDA outperforms the other classifiers. Algorithms have been run 50 times on each database and the average values together with the standard deviation (in parenthesis) of precision, recall, f-measure and NMI have been computed.

In Table \ref{tab1}, it is important to observe as GA-ICDA is able to obtain the perfect discrimination for the labels, by always obtaining the 1 value for the first test. Also, our technique outperforms, in most of the cases, the other two unsupervised classifiers (K-Means and Hierarchical) in terms of the performance measures. In fact, if we consider the f-measure, we observe that K-Means algorithm obtains a value of 0.6280 for Cyrillic, 0.6117 for angular Glagolitic and 0.6988 for round Glagolitic. Hierarchical clustering reaches a value of 0.5882 for Cyrillic, 0.5882 for angular Glagolitic and 0.5714 for round Glagolitic in terms of f-measure. If we consider the NMI measure, K-Means obtains a value of 0.4006, while hierarchical clustering has a value of 0.3198. GA-ICDA always obtains the best value of 1 for all the three script classes, in terms of both f-measure and NMI. Similarly can be discussed for precision and recall.

Results in Table \ref{tab2} confirm the superiority of the proposed approach with respect to the other two clustering algorithms. Looking at GA-ICDA results, it is interesting to observe that Cyrillic obtains a perfect distinction in terms of precision, recall and f-measure, reaching the maximum value of 1. GA-ICDA reaches a precision value of 0.8333 and a recall value of 1.0000 for distinction of angular Glagolitic and a precision value of 1.0000 and a recall value of 0.6000 for distinction of round Glagolitic. The NMI value is 0.7782. It is mainly due to the partial overlap between angular and round Glagolitic when the new labels in angular Glagolitic are added to the original set. This overlap in terms of classification is quite reasonable, because the new labels have originated in the late period of round Glagolitic, when the process of its conversion to angular Glagolitic was started. Consequently, the differences between the new set of labels in angular and round Glagolitic are really minimal. Table \ref{tab3} shows an example of the confusion matrix obtained from the classification of the labels with GA-ICDA when the new set of labels in angular Glagolitic is added. Overlap in classification is realized in terms of 2 labels in round Glagolitic classified as angular Glagolitic. Despite this consideration, it is important to observe as all the angular Glagolitic labels and most of the round Glagolitic labels (3 out of 5) are perfectly classified.

\begin{table}
\caption{Classification results in terms of precision, recall, f-measure and NMI obtained from GA-ICDA, K-Means and Average Linkage Hierarchical clustering on the first database of  labels in old Cyrillic, angular and round Glagolitic scripts.}
\begin{center}
\includegraphics[width=9cm,height=9cm,keepaspectratio]{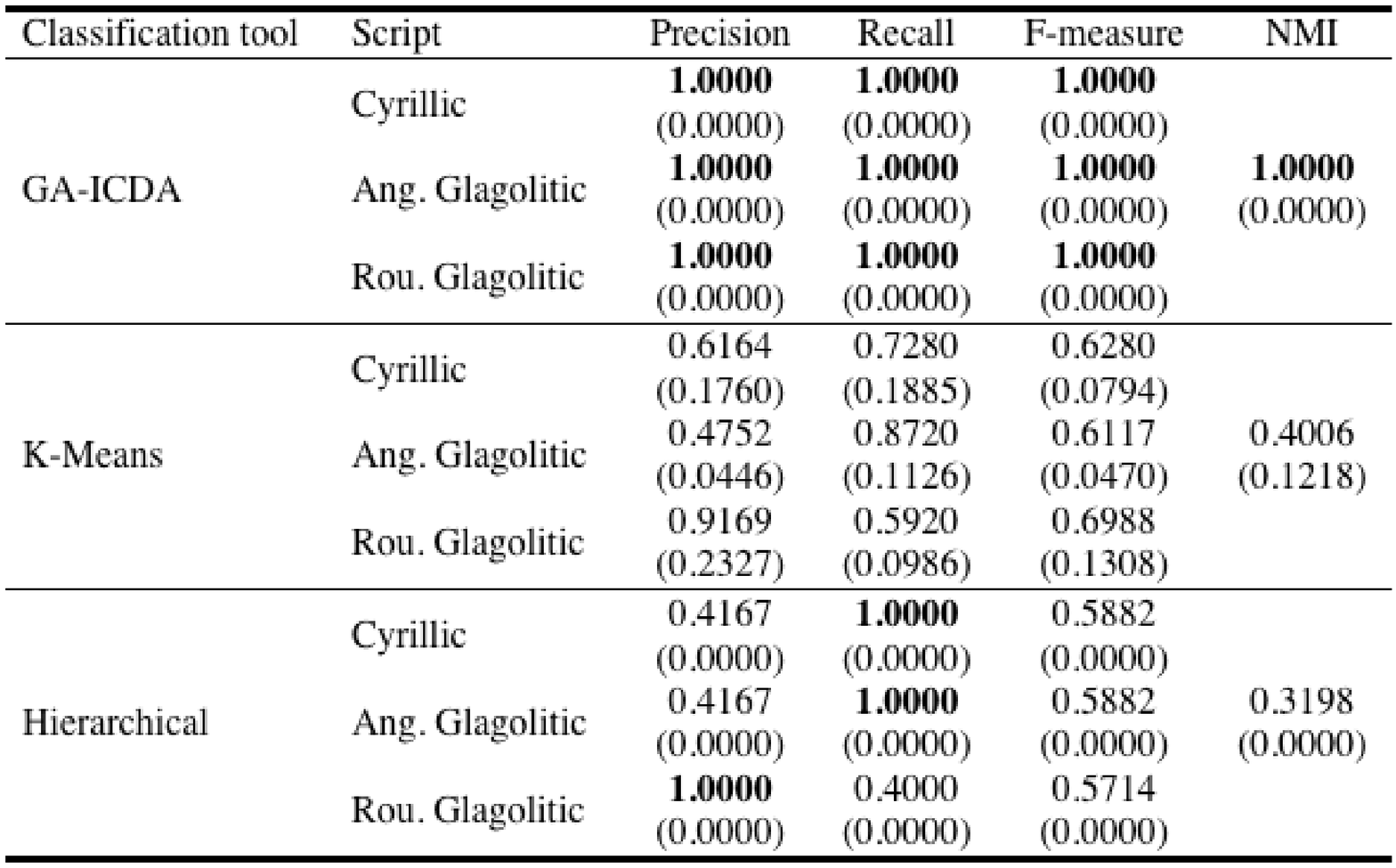}
\label{tab1}
\end{center}
\vspace{-0.6cm}
\end{table}

\begin{table}[!ht]
\caption{Classification results in terms of precision, recall, f-measure and NMI obtained from GA-ICDA, K-Means and Average Linkage Hierarchical clustering on the second database of labels in old Cyrillic, angular and round Glagolitic scripts.}
\begin{center}
\includegraphics[width=9cm,height=9cm,keepaspectratio]{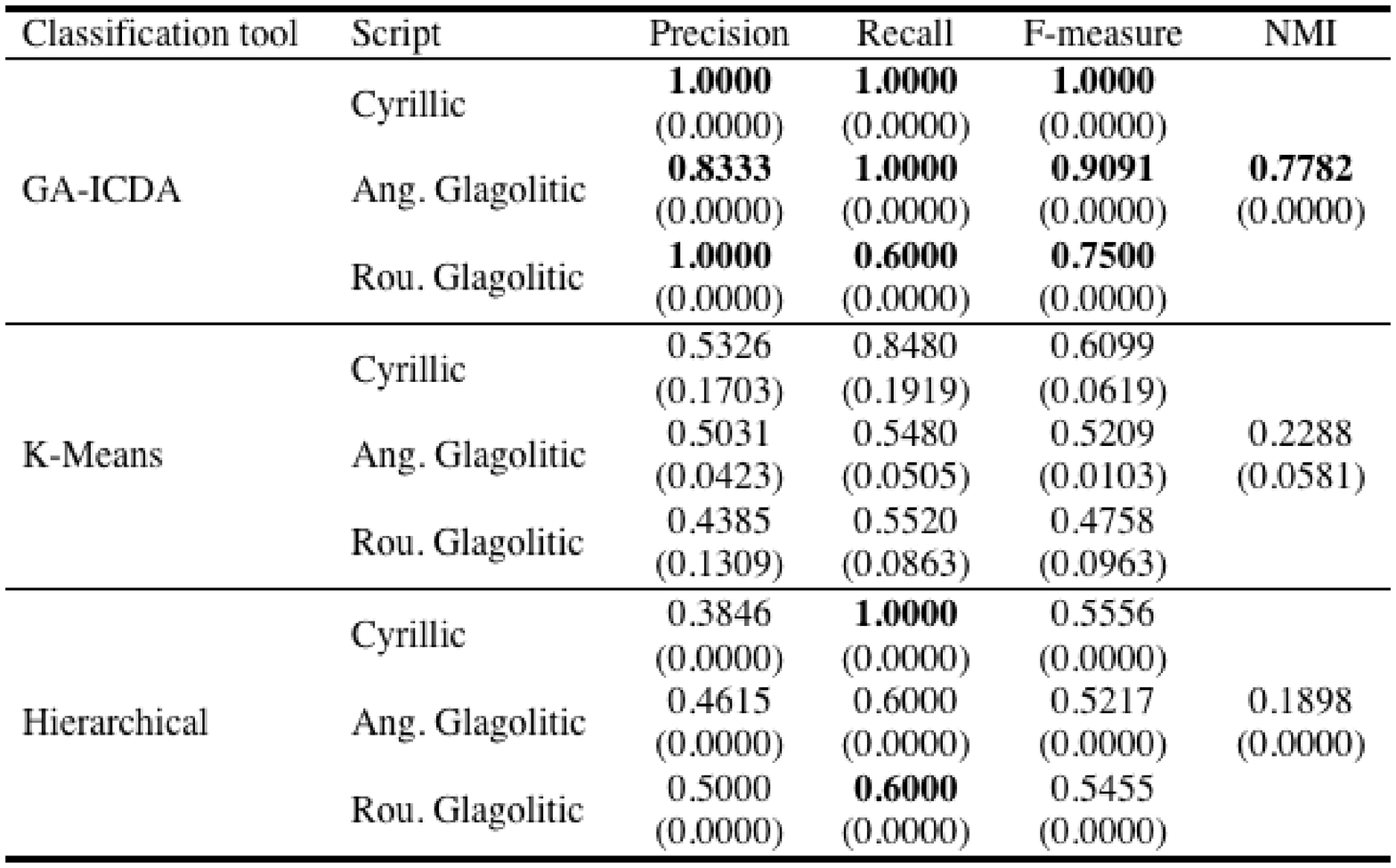}
\label{tab2}
\end{center}
\end{table}

K-Means and Hierarchical clustering don't reach the same results as GA-ICDA classifier. In fact, considering the f-measure, K-Means obtains a value of 0.6099 for the distinction of Cyrillic, a value of 0.5209 for the distinction of angular Glagolitic and a value of 0.4758 for the distinction of round Glagolitic. Hierarchical clustering reaches a value of 0.5556 for the distinction of Cyrillic, a value of 0.5217 for the distinction of angular Glagolitic and a value of 0.5455 for the distinction of round Glagolitic. GA-ICDA outperforms the other two classifiers by obtaining a value of 1.0000 for the distinction of Cyrillic, a value of 0.9091 for the distinction of angular Glagolitic and a value of 0.7500 for the distinction of round Glagolitic. In terms of NMI, K-Means algorithm reaches a value of 0.2288 and hierarchical clustering obtains a value of 0.1898, against a value of 0.7782 obtained from GA-ICDA.

\begin{table}[!ht]
\caption{Confusion matrix obtained from the GA-ICDA method for classification of the labels in Cyrillic, angular and round Glagolitic, when the new set of labels is added to the class of angular Glagolitic.}
\begin{center}
\includegraphics[width=6cm,height=6cm,keepaspectratio]{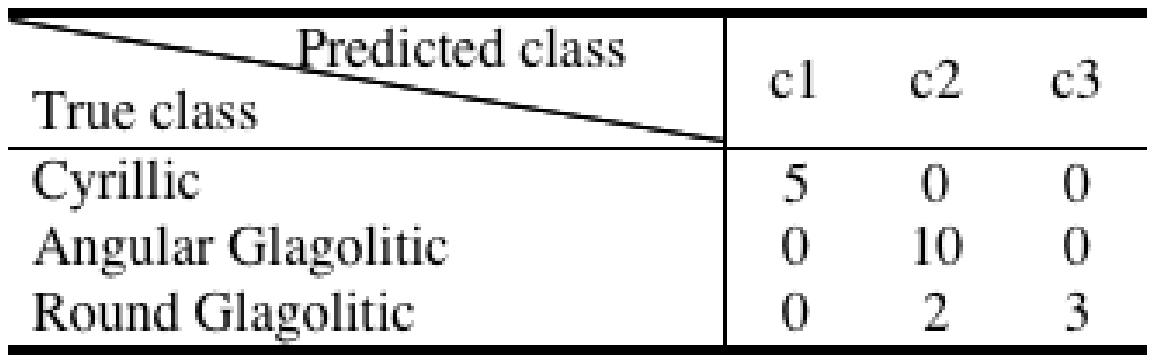}
\label{tab3}
\end{center}
\vspace{-0.5cm}
\end{table}

\section{Conclusion}
The paper proposed a new method for the script identification on the example of the old medieval labels written in South Slavic scripts: old Cyrillic, angular and round Glagolitic. The algorithm combines the run-length statistical analysis, and the adjacent local binary pattern analysis of the coded text obtained by mapping the initial text document (label). It shows meaningful discrimination between the three different script types. Hence, an extension of a state-of-the-art classification method is adopted for strong script classification and identification of the obtained statistical features. The proposed algorithm is evaluated on a custom-oriented database and on its extension to a more complex label set, which consist of labels written in old Cyrillic, angular and round Glagolitic scripts. The experiments gave very positive results.
Future work will extend the experimentation to larger datasets of labels.

%
% ---- Bibliography ----
%


\begin{thebibliography}{5}
%

\bibitem {Zhang2013}
Zhang, H., Zhao, K., Song, Yi-Zhe, Guo, J.:
Text extraction from natural scene image: A survey.
Neurocomputing 122(0), 310--323 (2013)

\bibitem {Fucic82}
Fu\v ci\'c, B.:
Glagoljski natpisi (in english "Glagolitic Labels"), Jugoslovenska akademija znanosti i umjetnosti.
Zagreb, (1982)

\bibitem {BrodicSC}
Brodi\'c, D., Milivojevi\'c, Z.N., Maluckov, \v C.A.:
An approach to the script discrimination in the Slavic documents.
Soft Computing, 1--11, In Press, DOI: 10.1007/s00500-014-1435-1

\bibitem {Brodic14}
Brodi\'c, D., Maluckov, \v C.A., Milivojevi\'c, Z.N., Draganov I.R.:
Differentiation of the script using adjacent local binary patterns.
AIMSA 2014. LNCS, vol. 8722, 162--169, Springer, Heidelberg (2014)

\bibitem {Preparata95}
Preparata, F. P., Shamos, M. I.:
Computational Geometry: An Introduction.
Springer, Berlin (1995).

\bibitem {Zram98}
Zramdini A.W., Ingold, R.:
Optical font recognition using typographical features.
IEEE Transaction on Pattern Analysis and Machine Intelligence 20(8), 877--882 (1998)

\bibitem {Tan98}
Tan, X.:
Texture information in run-length matrices.
IEEE Transaction on Image Processing 7(11), 1602--1609 (1998)

\bibitem{Galloway75}
Galloway, M.M.:
Texture analysis using gray level run lengths.
Computer, Graphics and Image Processing  4(2), 172--179 (1975)

\bibitem{Chu90}
Chu, A., Sehgal, C.M., Greenleaf, J.F.:
Use of gray value distribution of run lengths for texture analysis.
Pattern Recognition Letters 11(6), 415--419 (1990)

\bibitem{Dasar91}
Dasarathy, B.R., Holder, E.B.:
Image characterizations based on joint gray-level run-length distributions.
Pattern Recognition Letters, 12(8), 497--502 (1991)

\bibitem{Ojala96}
Ojala, T., PietikŠinen, M., Harwood, D.:
A comparative study of texture measures with classification based on feature distributions.
Pattern Recognition 29, 51--59 (1996)

\bibitem{Ojala2002}
Ojala, T., PietikŠinen, M., MŠenpŠŠ, T.:
Multi-resolution gray-scale and rotation invariant texture classification with local binary patterns.
IEEE Transaction on Pattern Analysis and Machine Intelligence 24, 971--987 (2002)

\bibitem{Amelio14}
Amelio, A., Pizzuti, C.:
A new evolutionary-based clustering framework for image databases.
ICISP 2014. LNCS, vol. 8509, 322--331, Springer International Publishing (2014)

\bibitem{Marti2001}
Marti, R., Laguna, M., Glover, F., Campos, V.:
Reducing the bandwidth of a sparse matrix with tabu search.
European Journal of Operational Research 135(2), 450--280 459 (2001)

\bibitem{Yuyu13}
Yuyu, Y., Xu, W., Yueming, L.:
A Hierarchical Method for Clustering Binary Text Image.
Trustworthy Computing and Services, CCIS 320, pp. 388--396, Springer (2013)

\bibitem{Aggar12}
Aggarwal, C. C., Zhai, C.:
A Survey of Text Clustering Algorithms.
In Aggarwal, C. C. et al (eds.), Mining Text Data, Springer US, pp. 77--128 (2012).

\bibitem{Zhu2006}
Zhu, Q., Li, J., Zhou, G., Li, P., Qian, P.:
A Novel Hierarchical Document Clustering Algorithm Based on a kNN Connection Graph.
ICCPOL 2006. LNAI, vol. 4285, 120--130, Springer, Heidelberg (2006)

\bibitem{Andrews2007}
Andrews, N.O., Fox, E.A.:
Recent Developments in Document Clustering.
Tech. rep., Computer Science, Virginia Tech (2007)

\bibitem{Vries12}
Vries, C.M.D., Geva, S., Trotman, A.:
Document clustering evaluation: Divergence from a random baseline.
CoRR abs/1208.5654 (2012)

\bibitem{Mann2009}
Manning, C.D., Raghavan, P., Schutze, H.:
Introduction to Information Retrieval.
Cam-bridge University Press, Online edition (2009)

\end{thebibliography}
\end{document}